\documentclass[9pt]{article}
\usepackage{spconf,amsmath,graphicx}
\usepackage{times}
\usepackage{epsfig}
\usepackage{algorithm}
\usepackage{algorithmic}
\usepackage{array}
\usepackage{cite}
\usepackage{enumerate}
\usepackage{multirow}
\usepackage{multicol}
\usepackage{arydshln}
\usepackage{color}
\usepackage{amssymb}
\usepackage{soul}
\usepackage{subfig}
\usepackage{float}
\usepackage{booktabs}
\usepackage{babel,blindtext}
\usepackage[normalem]{ulem}
\usepackage{microtype}

\def\0{{\mathbf 0}}
\def\1{{\mathbf 1}}

\def\a{{\mathbf a}}

\def\e{{\mathbf e}}

\def\v{{\mathbf v}}

\def\x{{\mathbf x}}
\def\y{{\mathbf y}}

\def\A{{\mathbf A}}
\def\B{{\mathbf B}}

\def\E{{\mathbf E}}

\def\G{{\mathbf G}}

\def\mH{{\mathbf H}}
\def\I{{\mathbf I}}
\def\J{{\mathbf J}}

\def\L{{\mathbf L}}

\def\R{{\mathbf R}}
\def\S{{\mathbf S}}

\def\U{{\mathbf U}}
\def\V{{\mathbf V}}
\def\W{{\mathbf W}}
\def\X{{\mathbf X}}

\def\ie{{\textit{i.e.}}}
\def\eg{{\textit{e.g.}}}

\def\cE{{\mathcal E}}

\def\cG{{\mathcal G}}
\def\cH{{\mathcal H}}

\def\cL{{\mathcal L}}

\def\cO{{\mathcal O}}

\def\cT{{\mathcal T}}

\def\cV{{\mathcal V}}
\def\cW{{\mathcal W}}

\def\bLambda{{\boldsymbol \Lambda}}

\def\diag{{\mathrm{diag}}}
\def\tr{{\mathrm{tr}}}

\title{Graph Learning with Spectral Connectivity Priors for Scarce Data}
\name{Mingxiao Liu$^\star$, Bahar Oveisgharan$^\dag$, Bingyan Zou$^\star$, Gene Cheung$^\dag$
, H. Vicky Zhao$^\star$, Feifei Gao$^\star$}
\address{$^\star$Tsinghua University, China ~~~~~~  $^\dag$York University, Canada 
}
\begin{document}
\ninept
\maketitle

\begin{abstract} 
Learning a sparse graph from scarce data is practically important but challenging.
Motivated by the desirable combination of local sparsity and strong global connectivity exhibited by expander-like graphs, we propose spectral connectivity-regularized graph learning (SCoGL), a framework that incorporates a family of Laplacian spectral priors to explicitly promote global connectivity.
Specifically, SCoGL augments a combinatorial-Laplacian-constrained graphical lasso (GLASSO) objective over a target adjacency matrix $\W$ with a general connectivity prior computed from Laplacian eigenvalues.
We derive gradients for several representative connectivity priors and develop a projected gradient descent (PGD) algorithm with Armijo backtracking to efficiently optimize $\W$.
Experiments show that the proposed SCoGL variants improve graph recovery and enhance downstream tasks such as graph signal denoising when signal observations are scarce. 
\end{abstract}

\section{Introduction}
\label{sec:intro}

A crucial prerequisite for \textit{graph signal processing} (GSP)
\cite{ortega18ieee,cheung18,Leus2023}---the study of mathematical tools such as transforms and wavelets that process discrete signals on finite graphs---is the construction of an underlying graph $\cG$ that encodes pairwise relationships appropriately. 
For a \textit{sparse} graph $\cG$ with $N$ nodes and $\cO(N)$ edges specified by Laplacian $\L \in \mathbb{R}^{N \times N}$, filtering operations such as $\L \x$ on graph signal $\x \in \mathbb{R}^N$ are linear time $\cO(N)$. 
Inferring a graph structure from observed data, called \textit{graph learning}, is extensively studied \cite{dong19,Kalofolias2016,Dong2016}.
Statistical approaches such as graphical lasso (GLASSO) \cite{Mazumder2012} and CLIME \cite{cai11_clime} are popular, and have been extended to specified Laplacian structures \cite{Egilmez2017,Zhao2019,Ying2020NGL,medvedovsky24,shi24,Yokota2025}. 
Geometric graphs based on feature distances learned from training data are also possible \cite{Hu2020,Yang2022}. 
Graphs can also be learned assuming a graph diffusion model over time \cite{Thanou2017}.

However, learning a sparse graph from scarce data, where the signal dimension $N$ far exceeds the number of observations $K$, \ie, $K \ll N$, is severely ill-posed and challenging. 
In such settings, additional structural priors are necessary to regularize graph estimation.
This regime is common in biological and biomedical applications, where high-dimensional measurements (\eg, gene expression, proteomic, or neuroimaging data) are often available for only a limited number of samples due to acquisition costs and experimental constraints \cite{Marbach2012}.

To address data scarcity, we introduce a family of global connectivity priors based on Laplacian eigenvalues to provide an inductive bias for graph estimation.
Because graph filtering is information diffusion among connected nodes, sparse graphs with strong global connectivity are desirable in GSP.
\textit{Expander graphs}\footnote{Expander graphs are sparse graphs with strong connectivity and no weak bottlenecks. Their expansion properties are closely related to the Laplacian spectral gap and, in particular, the algebraic connectivity $\lambda_2(\L)$ \cite{Hoory2006}.} \cite{Hoory2006} provide a canonical example of such structures.
Inspired by expander graphs, to promote global connectivity, we augment the classical GLASSO formulation with a general spectral connectivity prior computed from Laplacian eigenvalues, including the algebraic connectivity (Fiedler value, which is the second Laplacian eigenvalue $\lambda_2(\L)$) \cite{chung1997spectral} and the total effective resistance \cite{Klein1993}.

Prior work has incorporated connectivity into graph learning. Constraining a graph to be connected by imposing a positive lower bound on $\lambda_2(\L)$ was proposed in \cite{Sundin2017}, but the resulting semi-definite program (SDP) is computationally expensive.
Recently, \cite{Oveisgharan2026} proposed to augment GLASSO with a Fiedler value maximization term, $-\lambda_2(\L)$, for the scarce-data regime and greedily select one edge at a time for weight reduction or removal.
This heuristic edge-wise strategy does not jointly optimize all graph weights and requires many sequential updates, making it suboptimal and time-consuming.

In contrast, we propose \textit{spectral connectivity-regularized graph learning} (SCoGL), a framework that accommodates a family of global connectivity regularizers, where the Fiedler penalty is a special case. 
The resulting optimization problems are convex, allowing SCoGL to obtain globally optimal solutions, where the entire adjacency matrix $\W$ is treated as a single optimization variable.
We derive the gradients for several representative regularizers, including the Fiedler value~\cite{chung1997spectral} and total effective resistance~\cite{Klein1993}, and develop a \textit{projected gradient descent} (PGD) algorithm with Armijo backtracking~\cite{boyd04} that jointly updates all admissible edge weights at each iteration. 
Instead of sequential edge editing, these joint full-matrix updates enable efficient optimization.

Experimental results in the $K \ll N$ data regime show that the proposed SCoGL variants generally improve graph recovery in terms of relative error.
Moreover, when the learned graphs are used for downstream filtering tasks such as graph signal denoising, our proposed global connectivity prior enhances performance when signal observations are scarce. 

\section{Preliminaries}
\label{sec:prelim}

\subsection{GSP Definitions}
\label{subsec:gsp_def}

A positive graph $\cG(\cV,\cE,\W)$ is defined by a node set $\cV = \{1, \ldots, N\}$ and an edge set $\cE$, where $(i,j) \in \cE$ means nodes $i,j \in \cV$ are connected with positive weight $w_{i,j} = W_{i,j} \in \mathbb{R}_+$.
We assume edges are undirected, and thus \textit{adjacency matrix} $\W \in \mathbb{R}^{N \times N}$ is symmetric. 
The \textit{combinatorial graph Laplacian (CGL)} is defined as $\L \triangleq \mathrm{diag}(\W\1)-\W \in \mathbb R^{N\times N}. $ 
By construction, $\L\1=\0$, while $\L$ is real symmetric positive semidefinite (PSD) \cite{cheung18} and thus admits the eigen-decomposition $ \L=\V\bLambda\V^\top$ \cite{Horn2012}, 
where $\bLambda=\diag(\lambda_1,\ldots,\lambda_N)$ with $0=\lambda_1\leq\lambda_2\leq\cdots\leq\lambda_N$.
It is common in GSP to interpret the $k$-th eigenpair $(\lambda_k, \v_k)$ of $\L$ as the $k$-th graph frequency and graph Fourier mode for $\cG$. Let $\J\triangleq\frac{1}{N}\1\1^\top$ denote the orthogonal projector onto $\text{span}\{\1\}$.

\subsection{Global Connectivity Measures}
\label{subsec:connectivity-measures}

The combinatorial graph Laplacian's second smallest eigenvalue, $\lambda_2(\L)$, is the \textit{Fiedler value}, also known as algebraic connectivity~\cite{chung1997spectral,fiedler1973algebraic}.
It satisfies $\lambda_2(\L)\geq 0$, with equality if and only if the graph is disconnected. 
A unit-norm eigenvector $\v_2$ associated with $\lambda_2(\L)$ is called a \textit{Fiedler vector}. The eigen-pair $(\lambda_2,\v_2)$ characterizes the weakest nonconstant graph Fourier mode and indicates the presence of a global bottleneck of the graph through Cheeger-type relations~\cite{chung1997spectral}.

For a connected graph, the effective resistance between nodes $i$ and $j$ is $r_{ij}\triangleq(\e_i-\e_j)^\top\L^\dagger(\e_i-\e_j)$, where $\L^\dagger$ is the Moore--Penrose pseudoinverse of $\L$.
The \textit{total effective resistance} is
\begin{equation}
\label{eq:total-effective-resistance}
R_{\mathrm{tot}}(\L)
=N\tr(\L^\dagger)
=N\sum_{k=2}^{N}\frac{1}{\lambda_k(\L)}.
\end{equation}
It aggregates the entire nonzero Laplacian spectrum and is particularly sensitive to small eigenvalues. Accordingly, it has been proposed and studied as a measure of network robustness~\cite{Klein1993, ellens2011resistance}.

The two measures capture complementary aspects of global connectivity. The Fiedler value quantifies the weakest global bottleneck, whereas the total effective resistance aggregates resistance across all node pairs, reflecting path length, edge strength, and path redundancy.
As observed in~\cite{ellens2011resistance}, maximizing the Fiedler value and minimizing the total effective resistance can yield different optimal graphs.
We therefore consider both views in our framework.

\section{Problem Formulation}
\label{sec:formulate}

\subsection{Spectral Connectivity Regularizers}
\label{subsec:connectivity}

Building on the global connectivity measures reviewed in Section~\ref{subsec:connectivity-measures}, we construct connectivity regularizers using two spectral scopes: the \textit{Fiedler mode}, i.e., the \textit{weakest} nonconstant graph Fourier mode, characterized by $\lambda_2(\L)$, and the \textit{full spectrum}. The Fiedler-mode regularizer depends only on $\lambda_2(\L)$, whereas the full-spectrum regularizer aggregates all nonzero Laplacian eigenvalues.

For a connected graph Laplacian $\L$, we define
\begin{equation}
\label{eq:spectral-connectivity-family}
C_{\rho}^{\mathrm{weak}}(\L)
\triangleq \rho\bigl(\lambda_2(\L)\bigr)
\quad \text{and} \quad
C_{\rho}^{\mathrm{full}}(\L)
\triangleq \sum_{k=2}^{N}\rho\bigl(\lambda_k(\L)\bigr)
\end{equation}
as the Fiedler-mode and the full-spectrum regularizers, respectively. In \eqref{eq:spectral-connectivity-family}, $\rho:\mathbb{R}_{++}\rightarrow\mathbb{R}$ is a convex and non-increasing function. 
The non-increasing property makes minimization favor larger Laplacian eigenvalues, while convexity enables the resulting graph-learning problem to remain convex. 
As representative rather than exhaustive choices, we consider
\(
\rho_{\mathrm{lin}}(s)=-s,
\rho_{\log}(s)=-\log s,
\rho_{\mathrm{inv}}(s)=s^{-1}.
\)
The linear function provides a constant marginal reward for increasing spectral values, whereas the logarithmic and the inverse functions act as barrier-type penalties near zero with diminishing influence as $s$ increases. Combining the two spectrum scopes in~\eqref{eq:spectral-connectivity-family} with these three shaping functions yields the six representative regularizers summarized in Table~\ref{tab:connectivity-regularizers}. The suffixes in Table~\ref{tab:connectivity-regularizers} are used throughout the paper.

\begin{table}[t]
\centering
\caption{Representative spectral regularizers and corresponding
SCoGL suffixes (in parentheses).}
\label{tab:connectivity-regularizers}
\small
\setlength{\tabcolsep}{4.5pt}
\begin{tabular}{cccc}
\toprule
Spectral scope
& $\rho_{\mathrm{lin}}$
& $\rho_{\log}$
& $\rho_{\mathrm{inv}}$ \\
\midrule
\multirow{2}{*}{Fiedler mode}
& $-\lambda_2$
& $-\log\lambda_2$
& $\lambda_2^{-1}$ \\
& \textnormal{(F)}
& \textnormal{(LogF)}
& \textnormal{(InvF)} \\
\addlinespace[2pt]
\multirow{2}{*}{Full spectrum}
& \multirow{2}{*}{$-\tr(\L)$}
& $-\log\det(\L+\J)$
& $\tr(\L^\dagger)$ \\
& 
& \textnormal{(LogDet)}
& \textnormal{(R)} \\
\bottomrule
\end{tabular}
\end{table}

We next establish that regularizers of both spectrum scopes preserve convexity.

\smallskip
\noindent\textbf{Proposition 1.}
Let $\rho:\mathbb{R}_{++}\rightarrow\mathbb{R}$ be convex and
nonincreasing. Then both $C_{\rho}^{\mathrm{weak}}$ and
$C_{\rho}^{\mathrm{full}}$ are convex on the set of connected graph Laplacians.

\smallskip
\noindent\textit{Proof.}
Let $\U\in\mathbb{R}^{N\times(N-1)}$ have orthonormal columns spanning $\1^\perp$, and define the linear map $\B(\L)\triangleq\U^\top\L\U.$
For every connected graph Laplacian $\L$, we have $\B(\L)\succ0.$ 
Let $\boldsymbol{\lambda}(\B)\triangleq [\lambda_1(\B),\ldots,\lambda_{N-1}(\B)]^\top $, where $\{\lambda_k(\B)\}_{k=1}^{N-1}$ are eigenvalues of $\B$ in non-decreasing order.

For the Fiedler-mode regularizers, we have $\lambda_2(\L)=\lambda_{\min}\allowbreak\bigl(\B(\L)\bigr).$ The minimum eigenvalue is concave since $\lambda_{\min}(\B) =\min_{\|\x\|_2=1}\allowbreak\x^\top\B\x$ is the pointwise minimum of linear functions in $\B$.
Since $\B(\L)$ is linear in $\L$, $\lambda_2(\L)$ is concave in $\L$. Because $\rho$ is convex and nonincreasing, $C_{\rho}^{\mathrm{weak}}(\L)=\rho(\lambda_2(\L))$ is convex by the composition rule.

For the full-spectrum regularizers, define
$h(\x) \triangleq\sum_{i=1}^{N-1}\rho(x_i), \allowbreak\x\in\mathbb{R}_{++}^{N-1}.$
Because $\rho$ is convex, $h$ is convex; moreover, $h$ is invariant under permutations of its arguments.
From the standard convexity theorem for spectral functions~\cite{lewis1996convex}, $g(\B) \triangleq h\bigl(\boldsymbol{\lambda}(\B)\bigr) =\sum_{i=1}^{N-1}\rho\bigl(\lambda_i(\B)\bigr) $ is convex on the positive definite cone $\mathbb{S}_{++}^{N-1}$.
Consequently, $ C_{\rho}^{\mathrm{full}}(\L) =g\bigl(\U^\top\L\U\bigr) $ is convex because it is the composition of a convex function with a linear map.
\hfill$\square$



Proposition~1 shows that all six choices in Table~\ref{tab:connectivity-regularizers} define convex spectral regularizers on the set of connected graph Laplacians.
We next incorporate a spectral regularizer from this family into the graph-learning objective.

\subsection{Connectivity-Regularized Graph Learning Objective}
\label{subsec:graph-learning}

To formulate the graph learning objective, we define a \textit{Hilbert space} $\cH \subset \mathbb{R}^{N \times N}$ of real symmetric matrices endowed with the standard inner product $\langle \A, \B\rangle = \text{tr}(\A \B) = \sum_{i,j} A_{i,j} B_{i,j}$, for $\A, \B \in \cH$.
For positive and undirected graphs, 
define the \textit{convex cones} in $\cH$ of adjacency and CGL matrices as
\begin{equation}
\label{eq:graph-cones}
\begin{aligned}
\cW &\triangleq
\{\W\in\cH\mid W_{ij}\geq0,\ i\neq j,\ W_{ii}=0\}\\ \text{and} \quad
\cL &\triangleq
\{\L\in\cH\mid \L\1=\0,\ L_{ij}\leq0,\ i\neq j\},
\end{aligned}
\end{equation}
respectively.
Define the linear transformation $\cT:\cH\to\cH$ by $\cT(\W)\triangleq\diag(\W\1)-\W$. Since $\cT$ maps $\cW$ onto $\cL$, we parameterize $\L=\cT(\W)$.

To define the objective function, 
given $K$ zero-mean graph-signal observations, let $\X\triangleq[\x_1,\ldots,\x_K]\in\mathbb{R}^{N\times K}$ and
$\S\triangleq\X\X^\top/K$.
We augment the CGL-constrained GLASSO objective, which combines the Gaussian negative log-likelihood and $\ell_1$ penalty
\footnote{We also tested concave sparsity penalties as in~\cite{Ying2020NGL}. Under scarce data, they tended to produce a few excessively large edge weights, resulting in worse recovery performance than the $\ell_1$ penalty.} 
with a spectral connectivity regularizer:

\begin{equation}
\label{eq:main-objective}
\min_{\W \in \cW}
\underbrace{
\left\langle\S,\L\right\rangle
-\log\det(\L+\J)
}_{\text{data fidelity}} +
\underbrace{\mu\|\W\|_1}_{\text{sparsity}}
+
\underbrace{\gamma C(\L)}
_{\text{global connectivity}}
\end{equation}
where $\mu,\gamma\geq0$ control the sparsity and connectivity strengths, respectively. 
The function $C$ denotes the spectral regularizer defined in Section~\ref{subsec:connectivity}.
The constraint $\W\in\cW$ ensures that $\L=\cT(\W)\in\cL$ is a CGL.
By Proposition~1, all objective terms are convex on their effective domains, and the feasible set is convex. Therefore, \eqref{eq:main-objective} is a convex optimization problem. 

In Table~\ref{tab:connectivity-regularizers}, for $\W\in\cW$, the full-spectrum linear regularizer 
$C_{\mathrm{lin}}^{\mathrm{full}}(\L)=-\tr(\L)=-\|\W\|_1$ has the same form as the sparsity term. Substituting $C_{\mathrm{lin}}^{\mathrm{full}}(\L)$ into \eqref{eq:main-objective}, we have
\begin{equation}
\min_{\W\in\cW}
\quad
\left\langle\S,\L\right\rangle
-\log\det(\L+\J) +
(\mu-\gamma)\|\W\|_1.
\end{equation}
Thus, this regularizer defaults to the connectivity-free CGL-constrained GLASSO problem and is not considered as a separate
SCoGL variant.
In addition, the full-spectrum inverse regularizer $C_{\mathrm{inv}}^{\mathrm{full}}(\L)=\tr(\L^\dagger)=R_{\mathrm{tot}}(\L)/N$ is proportional to total effective resistance and is denoted by \textnormal{SCoGL-R}.

\subsection{Projected-Gradient Optimization}
\label{subsec:pgd}

We solve \eqref{eq:main-objective} using \textit{projected gradient descent} (PGD) with the adjacency matrix $\W$ as the optimization variable. This choice is motivated by the projection step: projection onto the CGL cone $\cL$ does not have a simple elementwise closed form, whereas projection onto the adjacency cone $\cW$ is available in closed form. Since the data-fidelity and spectral-regularization terms, together with their gradients, are naturally expressed in terms of the Laplacian $\L$, we use the adjoint of $\cT$ to map $\L$-domain gradients to the $\W$-domain.

The adjoint $\cT^*:\cH\rightarrow\cH$ is characterized by $\langle \cT(\W),\A\rangle = \langle \W,\cT^*(\A)\rangle $ and is given entrywise by $[\cT^*(\A)]_{ij}= \frac{1}{2}(A_{ii}+A_{jj})-A_{ij}.$
Hence, for a differentiable function $f$ of
$\L=\cT(\W)$, $\nabla_{\W}f(\cT(\W)) = \cT^*(\nabla_{\L}f(\L)).$

We next derive a unified gradient representation for the
spectral connectivity regularizers in Section~\ref{subsec:connectivity}.
For the Fiedler-mode regularizers, assuming $\lambda_2(\L)$ is simple with unit-norm eigenvector $\v_2$, we have
\begin{equation}
\label{eq:weak-gradient}
    \nabla_{\L} C_{\rho}^{\mathrm{weak}}(\L)
    =
    \rho'(\lambda_2)\v_2\v_2^\top .
\end{equation}
If $\lambda_2$ has multiplicity $m > 1$, $C_{\rho}^{\mathrm{weak}}$ is generally non-differentiable. 
A valid subgradient is $\rho'(\lambda_2)\E\R\E^\top$, where the columns of $\E\in\mathbb{R}^{N\times m}$ span the Fiedler eigenspace, $\R\succeq\0$, and $\tr(\R)=1$. The rank-one form in \eqref{eq:weak-gradient} is recovered by setting $\R=\a\a^\top$ and $\v_2=\E\a$ for any unit-norm vector $\a$.

For the full-spectrum regularizers, we have
\begin{equation}
\label{eq:full-gradient}
    \nabla_{\L} C_{\rho}^{\mathrm{full}}(\L)
    =
    \sum_{k=2}^{N}
    \rho'(\lambda_k)\v_k\v_k^\top .
\end{equation}

Equations~\eqref{eq:weak-gradient} and~\eqref{eq:full-gradient} provide the (sub)gradient of connectivity regularizers needed to form the objective update. At iteration $t$, let
$\L_t=\cT(\W^{(t)})$,
$\A_t=\L_t+\J$, $\E_{\mathrm{off}}\triangleq\1\1^\top-\I$, and
$\mH_t\in\partial_\L C(\L_t)$.
A (sub)gradient of the objective function in \eqref{eq:main-objective} is
\begin{equation}
\label{eq:objective-gradient}
    \G_t
    =
    \cT^*\!\left(
        \S-\A_t^{-1}+\gamma\mH_t
    \right)
    +\mu\E_{\mathrm{off}}.
\end{equation}

Here, $\cT^*(\S-\A_t^{-1})$, $\gamma\cT^*(\mH_t)$, and $\mu\E_{\mathrm{off}}$ are gradients of the data fidelity, connectivity, and sparsity terms, respectively. 
We update the entire matrix $\W$ via the PGD step
\begin{equation}
\label{eq:pgd-candidate}
    \widetilde{\W}_t(\eta)
    =
    \Pi_{\cW}
    \left(\W^{(t)}-\eta\G_t\right),
\end{equation}
where the projection $\Pi_{\cW}$ sets the diagonal to zero and clips negative off-diagonal entries to zero, thereby ensuring $\widetilde{\W}_t(\eta)\in\cW$, and $\eta > 0$ is the step size.
Armijo backtracking~\cite{boyd04} is used to select a step size $\eta_t > 0$ satisfying the sufficient-decrease condition, after which $\W^{(t+1)}=\widetilde{\W}_t(\eta_t)$.
Iterations terminate when the projected-gradient residual and
relative objective change fall below prescribed tolerances. 
The complete procedure of the proposed SCoGL algorithm is summarized in Algorithm~\ref{alg:scogl}.

\begin{algorithm}[t]
\caption{Spectral Connectivity-Regularized Graph Learning }
\label{alg:scogl}
\begin{algorithmic}[1]
\REQUIRE Sample covariance $\S$, parameters $\mu,\gamma$,
connectivity regularizer $C$, initialization $\W^{(0)}\in\cW$, max iteration $T_{\mathrm{PG}}$
\ENSURE Estimated adjacency matrix $\widehat{\W}$
\STATE $t\gets 0$

\WHILE{$t < T_{\mathrm{PG}}$ and not converged}
    \STATE $\L_t \gets \cT(\W^{(t)})$, \quad
           $\A_t \gets \L_t+\J$
    \STATE Compute $\mH_t\in\partial_\L C(\L_t)$
    \STATE Compute
    \(
        \G_t
        \gets
        \cT^*\!\left(
        \S-\A_t^{-1}+\gamma\mH_t
        \right)
        +\mu\E_{\mathrm{off}}
    \)
    \STATE Select $\eta_t$ by Armijo backtracking
    \STATE Update
    $\W^{(t+1)}
    \gets
    \Pi_{\cW}\!\left(
        \W^{(t)}-\eta_t\G_t
    \right),\quad t\gets t+1$
\ENDWHILE
\STATE \textbf{return} $\widehat{\W}\gets\W^{(t)}$
\end{algorithmic}
\end{algorithm}

\vspace{0.1in}
\noindent
\textbf{Complexity:}\,
Computing $\S$ requires $\mathcal{O}(KN^2)$ operations once before the iterations begin.
Each iteration is dominated by the factorization of $\A_t\in\mathbb{R}^{N\times N}$ and the required spectral computation, giving the worst-case complexity of $\mathcal{O}(N^3)$ in a dense implementation, while $\cT$, $\cT^*$, and $\Pi_{\cW}$ require only $\mathcal{O}(N^2)$ operations.
Since $K<N$, for $T_{\mathrm{PG}}$ iterations, the overall complexity is $\mathcal{O}(T_{\mathrm{PG}}N^3)$ with $\mathcal{O}(N^2)$ memory. The $\mathcal{O}(N^3)$ per-iteration complexity of SCoGL is of the same order as or lower than those of  \cite{Egilmez2017,Zhao2019,Ying2020NGL, Mazumder2012} that also address GLASSO-type objectives.

\section{Experiments}
\label{sec:experiments}

We evaluate SCoGL using weighted Erdős–Rényi graphs as a representative graph model and graph-signal denoising as a representative downstream task. 
We assess both graph recovery under different sample ratios and denoising performance in a severely undersampled setting.

\subsection{Experimental Setup}
\label{subsec:experimental-setup}

We test on weighted Erd\H{o}s--R\'enyi graphs with $N=100$ nodes and edge probability $p=0.15$. Nonzero edge weights are independently sampled from $\mathcal{U}[0.5,1.5]$. For each ground-truth Laplacian $\L_\star$, we generate $K$ i.i.d.\ intrinsic Gaussian graph signals $\x_k\sim\mathcal{N}(\mathbf{0},\L_\star^\dagger).$
To investigate the effect of sample scarcity, we consider sample ratio $K/N\in\{0.1,0.2,0.3,0.5,0.7,1.0\}$. 
The dataset contains 10 validation graphs and 10 test graphs. The validation graphs are used exclusively for objective parameter selection, after which the selected parameters are fixed and evaluated on the test graphs. At each sample ratio, all methods use the same graph and signal realizations.

We compare SCoGL against two general precision-matrix estimators,
GLASSO~\cite{Mazumder2012} and CLIME~\cite{cai11_clime}, as well as five graph-learning baselines: CGL-BCD~\cite{Egilmez2017}, GLENE~\cite{Zhao2019}, NGL~\cite{Ying2020NGL}, GL-SigRep~\cite{Dong2016} and Kalofolias~\cite{Kalofolias2016}. We evaluate the five connectivity-regularized instances retained from Section~\ref{subsec:graph-learning}: SCoGL-F, SCoGL-LogF, SCoGL-InvF, SCoGL-R, and SCoGL-LogDet.

To isolate the contribution of spectral connectivity regularization, we also include SCoGL-0 as a within-framework ablation. SCoGL-0 is obtained by setting $\gamma=0$ in~\eqref{eq:main-objective}. It retains the same CGL-constrained GLASSO objective and PGD solver as the five regularized variants, but removes the connectivity term. Consequently, comparisons between SCoGL-0 and SCoGL-F/LogF/InvF/R/LogDet measure the impact of the connectivity prior while holding the remaining model and optimization procedure fixed.


\subsection{Graph Recovery}
\label{subsec:recovery}

Our primary recovery metric is the relative Laplacian error $\mathrm{RE}\triangleq\|\widehat{\L}-\L_\star\|_F/\|\L_\star\|_F$, where $\widehat{\L}$ is the estimated Laplacian.

For each method and sample ratio, we perform an exhaustive grid search on the validation graphs to find the optimal parameter combination. We first discard parameter configurations whose mean edge density exceeds $0.25$, thereby restricting selection to sparse estimates, and then select the remaining configuration with the lowest mean validation RE. The selected parameters are fixed and evaluated on the 10 test graphs, and the mean test results are reported.

\begin{figure}[t]
    \centering
    \includegraphics[width=0.9\columnwidth]
        {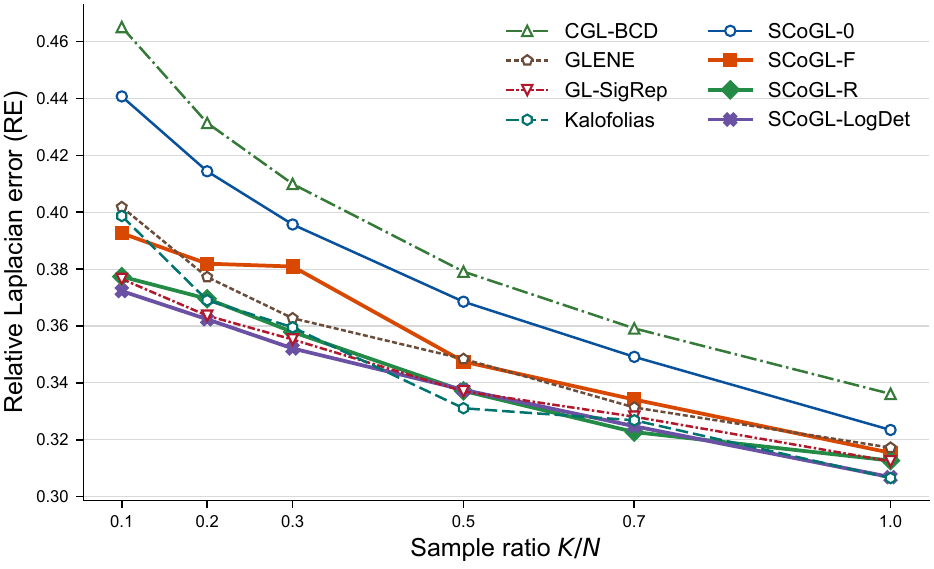}

    \caption{Mean test relative Laplacian error versus sample ratio $K/N$ for selected methods; lower is better.}
    \label{fig:recovery-k-sweep}
\end{figure}

\begin{table}[t]
\centering
\caption{Graph recovery and denoising results at $K/N=0.1$. Boldface indicates the best non-oracle result, and underlining indicates the second- and third-best results.}
\label{tab:main-results}
\footnotesize
\setlength{\tabcolsep}{5pt}
\renewcommand{\arraystretch}{1.10}
\begin{tabular}{cccc}
\toprule
\multirow{2}{*}{Method}
&
\multirow{2}{*}{RE $~\downarrow$}
&
\multicolumn{2}{c}{Denoising NMSE $~\downarrow$}
\\
\cmidrule(lr){3-4}
&
&
$\sigma=0.2$
&
$\sigma=0.4$
\\
\midrule
GLASSO~\cite{Mazumder2012}
    & 0.7185 & 0.3978 & 0.9209 \\
CLIME~\cite{cai11_clime}
    & 0.5718 & 0.3669 & 0.7558 \\
NGL~\cite{Ying2020NGL}
    & 0.9173 & 0.4703 & 1.4158 \\
CGL-BCD~\cite{Egilmez2017}
    & 0.4651 & 0.3552 & 0.7086 \\
GLENE~\cite{Zhao2019}
    & 0.4018 & 0.3482 & \underline{0.6871} \\
GL-SigRep~\cite{Dong2016}
    & \underline{0.3765} & 0.3488 & 0.6888 \\
Kalofolias~\cite{Kalofolias2016}
    & 0.3987 & 0.3502 & 0.6959 \\
SCoGL-0
    & 0.4407 & 0.3535 & 0.7102 \\
\midrule
SCoGL-F
    & 0.3926 & 0.3486 & 0.6917 \\
SCoGL-LogF
    & 0.3894 & \underline{0.3481} & 0.6905 \\
SCoGL-InvF
    & 0.3909 & 0.3486 & 0.6916 \\
SCoGL-R
    & \underline{0.3774} & \underline{0.3480} & \underline{0.6864} \\
SCoGL-LogDet
    & \textbf{0.3723}
    & \textbf{0.3474}
    & \textbf{0.6858} \\
\midrule
Oracle
    & 0.0000 & 0.3330 & 0.6552 \\
\bottomrule
\end{tabular}
\end{table}

Fig.~\ref{fig:recovery-k-sweep} compares RE across sample ratios $K/N$ for selected baselines and representative SCoGL variants for visual clarity. 
GLASSO, CLIME, and NGL are omitted because they yield substantially higher RE values. 
The second column of Table~\ref{tab:main-results} reports the RE of all methods in the most undersampled setting $K/N=0.1$. 
Across the evaluated sample ratios, CGL-BCD has the highest RE, while GL-SigRep, Kalofolias, SCoGL-R, and SCoGL-LogDet achieve comparatively small RE. 
At $K/N=0.1$, SCoGL-LogDet achieves the lowest RE of $0.3723$, followed by GL-SigRep at $0.3765$ and SCoGL-R at $0.3774$.

The comparison with SCoGL-0 further demonstrates the effectiveness of connectivity regularization. 
At $K/N=0.1$, SCoGL-0 obtains an RE of $0.4407$, whereas the five connectivity-regularized variants achieve RE values between $0.3723$ and $0.3926$, corresponding to reductions of $10.9\%$--$15.5\%$. 
Moreover, the three representative SCoGL variants plotted in Fig.~\ref{fig:recovery-k-sweep} consistently outperform SCoGL-0 across all evaluated sample ratios. These consistent gains demonstrate the effectiveness of the proposed spectral connectivity regularizers for graph recovery.

\subsection{Downstream Graph-Signal Denoising}
\label{subsec:denoising}

We next evaluate graph-signal denoising using the graphs learned in Section~\ref{subsec:recovery} under the most severely undersampled setting, $K/N=0.1$.
For each test graph, we generate $K_{\mathrm{denoise}}=150$ additional clean signals independently of those used for graph learning.
Noisy observations are generated as
$    \y
    =
    \x+\boldsymbol{\epsilon},
    \boldsymbol{\epsilon}
    \sim\mathcal{N}(\mathbf{0},\sigma^2\I).
$
We consider $\sigma\in\{0.2,0.4\}$. Given a learned Laplacian $\widehat{\L}$,  denoising is performed using the graph MAP filter
$
    \widehat{\x}
    =
    \left(\I+\sigma^2\widehat{\L}\right)^{-1}\y .
$

To assess the improvement toward the theoretical optimum, we also evaluate the above filter using the ground-truth Laplacian $\L_\star$, which we refer to as the oracle. The performance gap between SCoGL and the oracle quantifies the impact of graph estimation on the downstream denoising task.
Performance is measured by the aggregate normalized mean-squared error $\mathrm{NMSE}\triangleq\sum_k\|\widehat{\x}_k-\x_k\|_2^2/\sum_k\|\x_k\|_2^2$.

The last two columns of Table~\ref{tab:main-results} report the denoising results. SCoGL-LogDet attains the lowest non-oracle NMSE at both noise levels, achieving $0.3474$ at $\sigma=0.2$ and $0.6858$ at $\sigma=0.4$. SCoGL-R ranks second with NMSEs of $0.3480$ and $0.6864$, respectively. Both variants outperform all baselines, among which GLENE performs best with NMSEs of $0.3482$ and $0.6871$, respectively.

All five connectivity-regularized variants improve upon SCoGL-0 at both noise levels. In particular, SCoGL-LogDet reduces the NMSE of SCoGL-0 from $0.3535$ to $0.3474$ at $\sigma=0.2$ and from $0.7102$ to $0.6858$ at $\sigma=0.4$. 
These improvements close $29.8\%$ and $44.4\%$ of the corresponding SCoGL-0-to-oracle gaps. 
The performance gains at both noise levels indicate that spectral connectivity regularization provides a useful inductive bias for graph filtering when the graph is learned from scarce observations.

\section{Conclusion}
\label{sec:conclude}

In the data-scarce regime where the signal dimension $N$ far exceeds the number of observations $K$, \ie, $K \ll N$, reliable graph estimation requires additional structural priors. 
To this end, we proposed SCoGL, which augments a combinatorial-Laplacian-constrained GLASSO objective with a family of spectral connectivity priors derived from Laplacian eigenvalues, encompassing the Fiedler value and total effective resistance as special cases.
We derived the corresponding gradients and developed a projected gradient descent algorithm with Armijo backtracking that jointly optimizes the graph adjacency matrix $\W$.
Experimental results show that spectral connectivity regularization provides an effective inductive bias under severe data scarcity, yielding both more accurate graph recovery and improved graph signal denoising performance.

\appendix

\section*{\centering\normalsize ACKNOWLEDGMENT}
The work of G. Cheung was supported in part by the Natural Sciences and Engineering Research Council of Canada (NSERC) RGPIN-2025-06252.

\section*{\centering\normalsize COMPLIANCE WITH ETHICAL STANDARDS}
This is a numerical simulation study for which no ethical approval was required.


\begin{small}
\bibliographystyle{IEEEbib}
\bibliography{refs}
\end{small}

\end{document}